\def\0{\hbox{\bf 0\rm}}
\def\1{\hbox{\bf 1\rm}}

\def\Z{Z\kern -4pt Z}
\def \syntax {\circ_\triangleright}

\documentclass{article}
\usepackage{format/proceedings}
\usepackage{pstricks}
\usepackage{pst-node}

\newtheorem{theorem}{Theorem}

\newtheorem{definition}{Definition}
\newtheorem{proposition}{Proposition}

\newtheorem{example}{Example}

\title{Revising Partially Ordered Beliefs}

\author{ {\bf Salem Benferhat} \\  
  IRIT, Univerit{\'e} Paul Sabatier,\\
  118, route de Narbonne,\\
  31062 Toulouse Cedex 4, France\\ 
  benferhat@irit.fr \\
\And 
{\bf Sylvain Lagrue}  \\ 
PRAXITEC\\
  115, rue St Jacques,\\
  13006 Marseille, France\\ 
  lagrue@irit.fr\\
\And 
{\bf Odile Papini}   \\ 
  SIS, Universit{\'e} de Toulon et du Var,\\
  avenue de l'universit{\'e} BP 132,\\
  83957 La Garde cedex, France\\
  papini@univ-tln.fr\\
}

\begin{document}

\bibliographystyle{abbrv}
\maketitle

\begin{abstract}

This paper deals with the revision of partially ordered beliefs. It
proposes a semantic
representation of epistemic states by partial pre-orders on interpretations
and a syntactic representation by
partially ordered belief bases. Two revision operations, the revision
stemming from the history of observations and
the possibilistic revision, defined when the epistemic state is represented
by a total pre-order, are generalized, at
a semantic level, to the case of a partial pre-order on interpretations,
and at a syntactic level, to the case of
a partially ordered belief base. The equivalence between the two
representations is shown for the two revision operations.
\end{abstract}

%\newpage

\vspace{-0.25cm}

%%%%%%%%%%%%%%%%%%%%%%%%%%%
\section{Introduction}
%%%%%%%%%%%%%%%%%%%%%%%%%%%

%%%%%%%%%%%%%%%%%%%%%%%%
%\begin{itemize}
%\item[] introduction
%\item[] problematics
%\item[] semantic and syntactic revision of partial pre-orders
%\item[] from syntactic to semantic
%\item[] syntactic computation of $Bel(\Psi)$
%\item[] concluding discussion
%\end{itemize}
%%%%%%%%%%%%%%%%%%%%%%%%

Most of the time, an intelligent agent faces incomplete, uncertain or
inaccurate information. The arrival
of a new item of information,  more reliable or more certain leads the
agent to refine (specify)
 his beliefs, to revise them. Belief revision is a well known problem in
Artificial intelligence
\cite{AGM85}, \cite{PG88}, \cite{KM91}, in this context, an epistemic state
encodes a set of beliefs about the real world
(based on the available information). An epistemic state is generally
interpreted as a plausibility ordering
between possible states of the world, or as a preference relation between
information sources from
which an agent can derive his beliefs.

On a semantic level, epistemic states have been represented by a total
pre-order on interpretations
of the underlying logical language \cite{DP97}. This total pre-order models
the agent's preferences between several
situations. %Preferred situations are such that there are minimal according
%to the total-pre-order.
This pre-order has been encoded according to several ways, ordinals
\cite{WS87}, \cite{MAW94}, possibilities \cite{DuP92}, polynomials
\cite{OP01}.

However, %in case of partial ignorance because of incomplete information,
the agent has not always a total pre-order
between situations at his disposal, but is only able to define a partial
pre-order between situations.
The arrival of successive items of information can help him to refine this
partial pre-order in order
 to converge to a total pre-order between situations.

In other respects, total pre-orders are not suitable to model the case
where decision making is impossible or not arbitrary.
Suppose the agent has to make a decision on the cultivation of a new plot
according to three rules given by experts.
A first rule, $R1$ specifies that if some agronomical
conditions hold (warm climate, deep soil, acidity, etc...), called {\it
condition 1}, then
{\it the cultivation of tobacco is feasible}. The second rule $R2$
specifies that if the agronomical conditions ({\it condition 1}) hold and
the zone is mildewed (mildew parasite could ruin the plantation)
called {\it condition 2}, then {\it the cultivation of tobacco is not
feasible}. The third rule $R3$ specifies that according to the regulation
of production of
tobacco, if the area of the plot is not greater than the authorized area,
called {\it condition 3},
then {\it the cultivation of tobacco is feasible}.

The question is to define a pre-order on the three rules in order to make a
decision on the cultivation of tobacco.
Since $R2$ is more specific than $R1$, it is natural to prefer $R_2$ over $R_1$ (namely $R2 < R1$ holds). However, {\it
condition 3} is not related with {\it condition 1} nor with
{\it condition 2}, then it seems reasonable to consider $R3$ incomparable with
$R1$ and $R2$. On contrast, if we want
to impose a total pre-oder on rules, we have to set $R3$ relatively to
$R2$, therefore
there are two intuitive choices, either $R1  \leq R3$ or $ R3 \leq R2$. In the first
case, the following total pre-order holds:
$R2 < R1 \leq R3$ this means that if {\it condition 1}, {\it condition 2}
and {\it condition 3} are both satisfied,
according to the total pre-order we make the decision that {\it the
cultivation of tobacco is not feasible}. In
the second case, the following total pre-order holds: $ R3 \leq R2 < R1$
this means that if {\it condition 1}, {\it condition 2} and {\it condition
3} are satisfied,
according to the total pre-order we make the decision that {\it the
cultivation of tobacco is feasible}. These
two total pre-orders lead to contradictory decisions, and there is no
reason to choose the first one or the second one,
the choice can only be arbitrary. Since we think that arbitrary choices
have to be excluded, we think that a better
solution is to consider $R3$ and $R2$ as incomparable (and hence all total pre-orders are considered), and thus to define a
partial pre-order between rules. In such
cases, an epistemic state has to be represented by a partial pre-order on
interpretations and revision operations of
partial pre-orders by formulas have to be defined. Partial pre-orders on
interpretations have been used to represent
update operations \cite{KM91b}, this paper does not address updates but
revisions.

%In previous works, epistemic states have been represented by a total
%pre-order, and two kinds of representations
%have been proposed. At a semantic level, epistemic states have been
%represented by a total pre-order on interpretations (or worlds)
%and revision operations of a total pre-order by a formula have been defined
%\cite{BDP99}, \cite{OP01}. At a
%syntactic level, epistemic states have been represented by stratified
%belief bases and syntactic counterparts of semantic
%revision operations have been proposed \cite{DuP92}, \cite{BDP99}.
%Moreover, the equivalence between the two representations has
%been proved. 
In the present paper, we propose a generalization of two revision operators to the case of partial pre-orders.
%these two
%approaches representing epistemic states
%by either a partial pre-order on interpretations or a partially ordered
%belief base.
 Section $2$ presents the problematics of the representation of epistemic states by a
partial pre-order and focuses on the difficulties
of this generalization. Section $3$ presents the generalization of the
semantic revision operations to partial pre-orders and
the generalization of the syntactic counterparts of these operations to
partially ordered belief bases. These generalizations
are rather direct. On contrast the generalization of the mapping from the
syntactic level to the
semantic level is more problematical as shown in Section $4$, because the
definition of a partial pre-order between formulas
leads to two possible partial pre-orders between subsets of formulas. We
choose one of them, however the results presented hold
for the other one. We finally present the syntactic computation of the
belief set corresponding to an epistemic state
in Section $5$ before a concluding discussion in Section $6$.

%%%%%%%%%%%%%%%%%%%%%%%%%%%
\section{Presentation of the problem}
%%%%%%%%%%%%%%%%%%%%%%%%%%%

\subsection{Basic definitions of partial pre-orders}

In this paper we use propositional calculus, denoted by ${\cal L_{PC}}$, as
knowledge representation language with usual
connectives $\lnot$, $\land$, $\lor$, $\to$, $\equiv$ (logical equivalence).
The lower case letters $a$, $b$, $c$, $\cdots$, are used to denote
propositional variables,
lower case Greeks letters $\phi$, $\psi$, $\cdots$, are used to denote
formulas,
upper case letters $A$, $B$, $C$, are used to denote sets of formulas, and
upper case Greeks letters
$\Psi$, $\Phi$ $\cdots$, are used to denote epistemic states.
We denote by ${\cal W}$ the set of interpretations of ${\cal L_{PC}}$ and
by $Mod(\psi)$ the set of models of a formula $\psi$, that is
$Mod(\psi) = \{ \omega \in {\cal W}, \omega \models \psi \}$ where
$\models$ denotes the inference relation used for
drawing conclusions. %The symbol $\equiv$ denotes the logical equivalence.
\bigskip

%Generally, epistemic states are represented by total pre-orders on
%interpretations, however, as mentioned in the introduction,
%in case of partial ignorance, or decision making, the agent is unable to
%compare all situations between them and a partial
%pre-order seems to be more suitable to represent epistemic states.
%\bigskip

A partial pre-order, denoted by $\preceq$ on a set $A$ is a reflexive and
transitive binary relation.
Let $x$ and $y$ be two members of $A$, the equality is defined by $x = y$
iff  $x \preceq y$ and $y \preceq x$.
The corresponding strict partial pre-order, denoted by $\prec$, is such that,
$x \prec y$ iff $x \preceq  y$ holds but $x \preceq y$ does not hold.
 We denote by $\sim$ the incomparability relation $x \sim y$ iff $x \preceq
y$ does not hold nor $y \preceq x$.

$\preceq$ can equivalently be defined from $=$ and $\prec$: $a\preceq b$ iff $a\prec b$ or $a=b$. %This method shall be used in the rest of this paper, in order to provide  ``natural'' properties to $\preceq$.

Given $\preceq$ on a set $A$, the minimal elements of $A$ are defined by: 
$min(A,\preceq)=\{x:x\in A,\not \exists y\in A,y\prec x\}$

%In the mapping from the syntactic aspect to the semantic aspect of the
%representation of an epistemic state by a partial
%pre-order, 
For the purpose of this paper, we need to define a partial pre-order on subsets of elements of a set $A$.
According to Halpern's works \cite{JH97}, \cite{JH01} (see also Cayrol et al. \cite{crs1992}, Dubois et al. \cite{dlp1992} and Lafage et al. \cite{lls1999}) there are two
meaningful ways to compare subsets of $A$. We denote these two partial pre-orders on $2^{A}$ by $\preceq_{A,w}
$ and by $\preceq_{A,s}$.

Let $X$ and $Y$ be two subsets of $A$, %we denote by $min(X,
%\preceq_{A})$ resp. by $min(Y, \preceq_{A})$
%the preferred elements of $X$, resp. of $Y$. 
The sets $X$ and $Y$ are
considered equals if for each
preferred element in $X$ there exists a so preferred element in 
%the preferred elements of 
$Y$ and conversely, more formally:

\begin{definition}
Let $X$ and $Y$ be two subsets of $A$ and $\preceq_{A} $ a
partial pre-order on $A$, $X = Y$ iff  : \\

\begin{tabular}{c}
  $\forall x \in min(X, \preceq_{A}), \; \exists y \in
min(Y, \preceq_{A})$ such that
$x = y$ \\ {and} \\
  $\forall y \in min(Y, \preceq_{A}), \; \exists x \in min(X,
\preceq_{A})$ such that
$x=y$.
\end{tabular}
\end{definition}

The first way to define a partial pre-order on subsets $X$ and $Y$ of
$A$ is to consider that $X$ is preferred
to $Y$ if for each element of $ Y$ there exists at least one element in
$X$, which is preferred to it, more
formally (we assume that $X$ and $Y$ are not both empty):

\begin{definition}
  \label{def1}
%  Let $\preceq_A$ be a partial pre-order on $A$, $X,Y\subseteq A$ and assume that $X$ and $Y$  are not simultaneously empty. 
$X$ is weakly preferred to $Y$, denoted by $X\prec_{A,w}Y$, iff $\forall y\in Min(Y,\preceq_A)$, $\exists x\in Min(X,\preceq_A)$ such that $x\prec_A y$.
\end{definition}

%\begin{definition}
%\label{def1}
%Let $X$ and $Y$ be two subsets of $A$ and $\preceq_{A} $ a
%partial pre-order on $A$,
%$X \preceq_{A,w} Y$ iff $\forall y \in min(Y, \preceq_{A}) \;
%\exists x \in min(X, \preceq_{A}) $
%such that $x \preceq_{A} y$.

%\end{definition}

The second way to define a partial pre-order on subsets $X$ and $Y$ of
$A$ is to consider that $X$ is preferred
to $Y$ if there exists at least one element in $X$ which is preferred to
all elements in $Y$, more
formally:

\begin{definition}
  \label{def2}
  Let $\preceq_A$ be a partial pre-order on $A$ and $X,Y\subseteq A$. $X$ is strongly preferred to $Y$, denoted by $X\prec_{A,s}Y$ iff $\exists x\in Min(X,\preceq_A)$ such that $\forall y\in Min(Y,\preceq_A),$  $x\prec_A y$.
\end{definition}

%\begin{definition}
%\label{def2}
%Let $X$ and $Y$ be two subsets of $A$ and $\preceq_{A} $ a
%partial pre-order on $A$,
%$X \preceq_{A,s} Y$ iff $\exists x \in min(X, \preceq_{A})$ such
%that $\forall y \in min(Y, \preceq_{A})$,
%$ x \preceq_{A} y$.
%\end{definition}

It can be shown that the definition of $\prec_{A,s}$ implies the
definition of $\prec_{A,w}$, namely, if $X \prec_{A,s} Y$ then
$X \prec_{A,w} Y$. The converse does not hold. 

\begin{example}
Let $A = \{x_1, x_2, y_1, y_2  \}$ and $\preceq_{A}$ be a partial
pre-order on $A$ such
that $x_1 \preceq_{A} y_1$ and  $x_2 \preceq_{A} y_2$. Let $X$
and $Y$ be two subsets of $A$, $X =\{x_1, x_2 \}$
and $Y = \{y_1, y_2  \}$, we have $X \prec_{A,w} Y$, indeed $x_1$ is
preferred to $y_1$ and $x_2$ is preferred to
$y_2$. However, $X\prec_{A,s} Y$ does not hold, there is no element in $X$ which is preferred to all elements
of $Y$.
\end{example}

In the case where $\preceq_{A}$ is a total
pre-order, the definition of $\prec_{A,s}$ is equivalent to the
definition of $\prec_{A,w}$,
more formally, $X \prec_{A,s} Y$ iff $X \prec_{A,w} Y$.

%\begin{example}
%Let $A = \{x_1, y_1, y_2  \}$ and $\preceq_{A}$ be a partial
%pre-order on $A$ such
%that $x_1 \preceq_{A} y_1$ and  $x_1 \preceq_{A} y_2$. Let $X$
%and $Y$ be two subsets of $A$, $X =\{x_1, x_2 \}$
%and $Y = \{y_1, y_2  \}$, we have $X \prec_{A,w} Y$, and $X
%\prec_{A,s} Y$ since there
%exists an element in $X$, $x_1$,  which is preferred to all elements
%of $Y$.
%\end{example}

For lake of space, we will only focus on the weakly preference definition. But all provided results are valid for the strong preference.

\subsection{Semantic representation of epistemic states}

Let $\Psi$ be an epistemic state, $\Psi$ is first represented by a partial
pre-order on interpretations, denoted by
$\preceq_{\Psi}$. 
%This partial pre-order represents the agent's partial
%preferences between interpretations. 
The interpretation $\omega$ is preferred (or more plausible than) to $\omega'$,
denoted $\omega \preceq_{\Psi} \omega'$. 
$\omega  \sim_{\Psi} \omega'$ denotes that the agent has no preference
between $\omega$ and $\omega'$. The belief
set corresponding to $\Psi$, denoted by $Bel^{se}(\Psi)$, modeling the
agent's current beliefs is such that
$Mod(Bel^{se}(\Psi)) = min({\cal W}, \preceq_{\Psi})$. We illustrate this
representation with the example informally described in
the introduction.

\begin{example}
\label{ex1}

The agent has to make a decision on the cultivation of a new plot according
tree rules given by experts. A first rule, $R1$ specifies that if some
agronomical
conditions hold (warm climate, deep soil, acidity, etc...), called {\it
condition 1}, then
{\it the cultivation of tobacco is feasible}. The second rule $R2$
specifies that if the agronomical conditions ({\it condition 1}) hold and
the zone is mildewed (mildew parasite could ruin the plantation)
called {\it condition 2}, then {\it the cultivation of tobacco is not
feasible}. The third rule $R3$ specifies that
according to the regulation of the production of
tobacco, if the area of the plot is not greater than the authorized area,
called {\it condition 3},
then {\it the cultivation of tobacco is feasible}. More formally, the three
rules can be represented as follows:
$R1$: $b \to a$, $R2$: $b \land c \to \lnot a$, $R3$: $d
\to a$, where $a$ encodes
{\it the cultivation of tobacco is feasible}, $b$ encodes {\it condition
1}, $c$ encodes {\it condition 2}
and $d$ encodes {\it condition 3}.

There are four propositional variables $a$, $b$, $c$ and $d$. The sixteen
interpretations are $\omega_0 = \{\lnot a, \lnot b, \lnot c, \lnot d \}$,
$\omega_1 = \{\lnot a, \lnot b, \lnot c, d \}$, $\omega_2 = \{\lnot a, \lnot b,
c, \lnot d \} $, $\cdots$, $\omega_{14} =\{ a, b, c, \lnot d\}$,
$\omega_{15} = \{a, b, c, d \}$. Let $\Psi$ be an epistemic state which
corresponding belief set is
$Bel^{se}(\Psi) = (b \to a) \land (b \land c \to \lnot a) \land (d \to a)$, we represent the epistemic
state by a partial pre-order, denoted $\preceq_{\Psi} $ as follows:

%The interpretations that satisfies $R1$, $R2$ and $R3$ are preferred and
%$Mod(Bel^{se}(\Psi)) = \{\omega_0, \omega_2, \omega_8, $
%$\omega_9, \omega_{10}, \omega_{11}, \omega_{12}, \omega_{13} \}$. 
Since $R2$ is more specific than $R1$, and since $R2$ and $R3$ are
incomparable, then: %the intepretations that satisfy $R2$ and $R3$ but not $R1$,
%the ones that satisfy $R2$ and $R1$ but not $R3$ and
%the ones that satisfy $R2$ but not $R1$ but not $R3$ are then preferred,
%there are incomparable between them. 
\begin{itemize}
\item
  the interpretations satisfying all constraints are preferred to all other interpretations,
\item
  the interpretations which falsify $R_2$ are preferred to the interpretations falsifying $R_1$,
\item
  the interpretations which falsify $R_2$ and the ones which falsify $R_3$ are uncomparable.
\end{itemize}

The partial pre-order $\preceq_\Psi$ is represented  by Figure~\ref{initial} (an arrow $x\longleftarrow y$ means $x\prec y$, the transitivity and the reflexivity are not represented for sake of clarity).

\begin{figure}[h]
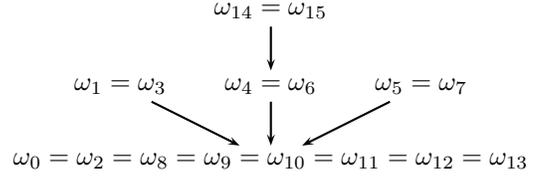

  \caption{Representation of initial epistemic state $\Psi$}\label{initial}
\vspace{2cm}
  \centering
%  \psframebox{
                                %    \psorigin(0,0)
    \rput(0,0){\rnode{I5}{$\omega_0=\omega_2=\omega_8=\omega_9=\omega_{10}=\omega_{11}=\omega_{12}=\omega_{13}$}}
    \rput(-2,1){\rnode{I1}{$\omega_1=\omega_3$}}
    \rput(0,1){\rnode{I2}{$\omega_4=\omega_6$}}
    \rput(2,1){\rnode{I3}{$\omega_5=\omega_7$}}
    \rput(0,2){\rnode{I4}{$\omega_{14}=\omega_{15}$}}
    \ncline[nodesep=3pt]{<-}{I5}{I1}
    \ncline[nodesep=3pt]{<-}{I5}{I2}
    \ncline[nodesep=3pt]{<-}{I5}{I3}
    \ncline[nodesep=3pt]{<-}{I2}{I4}
%}

\end{figure}

\end{example}

%%%%%%%%%%%%%%
%$$
%\begin{array}{ccc}
%& &  \\
%\omega_{14} =_{\Psi} \omega_{15} & &\\
%& &  \\
%\downarrow & &  \\
%& &  \\
% \omega_4 =_{\Psi}  \omega_6  & \omega_1 =_{\Psi} \omega_3 & \omega_5
%=_{\Psi} \omega_7 \\
%& &  \\
%\downarrow \quad & \downarrow \quad & \downarrow \\
%& &  \\
% & \omega_0 =_{\Psi} \omega_2 =_{\Psi} \omega_8 =_{\Psi} \omega_9 =_{\Psi}
%\omega_{10} =_{\Psi} \omega_{11} =_{\Psi}
%\omega_{12} =_{\Psi} \omega_{13} & \\
%\end{array}
%$$\\
%%%%%%%%%%%%%%

%\noindent $\omega_0 =_{\Psi} \omega_2 =_{\Psi} \omega_8 =_{\Psi} \omega_9
%=_{\Psi} \omega_{10} =_{\Psi} \omega_{11} =_{\Psi} \omega_{12} =_{\Psi}
%\omega_{13}
%\prec_{\Psi} \omega_4 =_{\Psi}  \omega_6
%\prec_{\Psi} \omega_{14} =_{\Psi} \omega_{15}$ and

%\noindent $\omega_0 =_{\Psi} \omega_2 =_{\Psi} \omega_8 =_{\Psi} \omega_9
%=_{\Psi} \omega_{10} =_{\Psi} \omega_{11} =_{\Psi}
%\omega_{12} =_{\Psi} \omega_{13}
%\prec_{\Psi} \omega_1 =_{\Psi} \omega_3$ and

%\noindent $\omega_0 =_{\Psi} \omega_2 =_{\Psi} \omega_8 =_{\Psi} \omega_9
%=_{\Psi} \omega_{10} =_{\Psi} \omega_{11} =_{\Psi}
%\omega_{12} =_{\Psi} \omega_{13}
%\prec_{\Psi} \omega_5 =_{\Psi} \omega_7$ and

%\noindent $ (\omega_{14} =_{\Psi} \omega_{15}) \sim_{\Psi} (\omega_1
%=_{\Psi} \omega_3)$,
%$(\omega_{14} =_{\Psi} \omega_{15}) \sim_{\Psi} (\omega_5 =_{\Psi} \omega_7 )$,
%$(\omega_4 =_{\Psi}  \omega_6) \sim_{\Psi} (\omega_1 =_{\Psi} \omega_3)$,
%$(\omega_4 =_{\Psi}  \omega_6) \sim_{\Psi} (\omega_5 =_{\Psi} \omega_7 )$,
%$(\omega_1 =_{\Psi}  \omega_3) \sim_{\Psi} (\omega_5 =_{\Psi} \omega_7 )$.

%\end{example}

\subsection{Syntactic representation of epistemic states}

An epistemic state $\Psi$ is syntactically represented by a
partially ordered belief base,
denoted by $\preceq_{\Sigma}$, where $\Sigma$ is a set of propositional
formulas, and $\preceq_{\Sigma}$ is a partial
pre-order on $\Sigma$. Let $\phi$ and $\phi'$ $\in \Sigma$,
 $\phi \preceq_{\Sigma} \phi'$ means that $\phi$ is preferred (more
important than) to $\phi'$ and $\phi
\sim_{\Sigma} \phi'$ denotes that the agent has no preference between
$\phi$ and $\phi'$.
We illustrate this representation with the example informally described in
the introduction.

\begin{example}
\label{ex2}

Let $\Psi$ be the epistemic state, where
$\Sigma = \{ b \to a, b \land c \to \lnot a, d \to
a \}$, we represent the epistemic
state by a partial pre-order on $\Sigma$, denoted by $\preceq_{\Sigma}, $ as
follows:
Since $ b \land c \to \lnot a$ is more specific than $b \to
a$, and $d \to a$ and
$ b \land c \to \lnot a$ are incomparable,
%%%%%%%%%%%%%%%%
%$$
%\begin{array}{cc}
%&   \\
% b \to a
%&   \\
%\downarrow &\\
%&   \\
% b \land c \to \lnot a & \qquad d \to a\\
%\end{array}
%$$
%%%%%%%%%%%%%%%%
the following partial pre-order on formulas holds:

\noindent $  b \land c \to \lnot a \prec_{\Sigma}  b \to a$
and $b \land c \to \lnot a \sim_{\Sigma} d \to a$
and $  b \to a \sim_{\Sigma} d \to a$.

\end{example}

The generalization of the representation and revision of an epistemic state to a partial
pre-order leads to the following diagram:

$$
\begin{array}{ccc}
\preceq_{\Sigma}  & \to & \preceq_{\Psi} \\
 \downarrow & & \downarrow \\
\preceq_{\Sigma \circ^{sy} \mu}  & \to & \preceq_{\Psi \circ^{se}\mu} \\
 \downarrow & & \downarrow \\
Bel^{sy}(\Psi \circ^{sy} \mu) & \equiv & Bel^{se}(\Psi \circ^{se} \mu)
\end{array}
$$

In the special case of a total pre-order, the diagram has been shown valid
and the equivalence between the syntactic
approach and the semantic approach has been proved \cite{BDP99}. The
question is what does remain true when this diagram is extended to the
representation
by a partial pre-order. We show in Section $3$ that we get the mappings
$\preceq_{\Sigma} \to \preceq_{\Sigma \circ^{sy} \mu}$
, $\preceq_{\Psi} \to \preceq_{\Psi \circ^{se} \mu}$ and
$\preceq_{\Psi \circ^{se} \mu} \to Bel^{se}(\Psi \circ^{se} \mu)$
rather directly. On contrast, the mappings $\preceq_{\Sigma}  \to
\preceq_{\Psi}$ and
$\preceq_{\Sigma \circ^{sy} \mu} \to Bel^{sy}(\Psi \circ^{sy} \mu)$
are less straightforward.%more problematical,
%because there are not unique.

We now present the revision extended to a partial pre-order.

%%%%%%%%%%%%%%%%%%%%%%%%%%%
\section{Semantic and syntactic revision of partial pre-orders}
%%%%%%%%%%%%%%%%%%%%%%%%%%%

\subsection{Extension of the revision stemming from the history of
observations}

We extend the revision operation, defined in \cite{BDP99, OP01}, to the case of partial pre-orders. The underlying intuition stems from the fact that the agent remembers all his previous observations. However these observations are not at the same level, according to whether there are more plausible or not in the next epistemic state. The general philosophy is that an old assertion is less reliable than a new one. In prediction problems, it seems reasonable to decrease the confidence that one has in an item of information, as time goes by. However, this revision operation attempts to satisfy as many previous observations as possible. That is, an old observation persists until it becomes contradictory with a more recent one. The revision operation uses the history of the sequence of previous observations to perform revision.

\subsubsection{Semantic extension}

When an epistemic state, $\Psi$, is represented by a partial pre-order on interpretations, the revision of $\Psi$ by a propositional formula $\mu$ leads to a revised epistemic state $\Psi \circ^{se}_{\triangleright} \mu$, represented by a partial pre-order on interpretations. This new epistemic state is such that the relative ordering between models of $\mu$ is preserved, the relative ordering between counter-models of $\mu$ is preserved, and the models of $\mu$ are preferred to its counter-models. More formally:

\begin{definition}\label{defsemo}
Let $\Psi$ be an epistemic state and $\mu$ be a propositional formula, the revised epistemic state $\Psi \circ^{se}_{\triangleright} \mu$ corresponds to the following partial pre-order:
\begin{itemize}
\item if $\omega, \, \omega' \, \in Mod(\mu)$ then $\omega \preceq_{\Psi\circ^{se}_{\triangleright} \mu} \omega'$
iff $\omega \preceq_{\Psi} \omega'$,
\item if $\omega, \, \omega' \, \not\in Mod(\mu)$ then $\omega
\preceq_{\Psi \circ^{se}_{\triangleright} \mu} \omega'$
iff $\omega \preceq_{\Psi} \omega'$,
\item if $\omega \in Mod(\mu)$ and $ \omega' \not\in Mod(\mu)$ then $\omega
\prec_{\Psi \circ^{se}_{\triangleright} \mu} \omega'$.
\end{itemize}
\end{definition}

\noindent According to this definition it is easy to check that
$Mod(Bel^{se}(\Psi  \circ^{se}_{\triangleright} \mu)) = min(Mod(\mu), \preceq_{\Psi})$.

\begin{example}
\label{ex5}
We come back to example~\ref{ex1}, where the initial epistemic state $\Psi$ is represented by the Figure~\ref{initial}.

\noindent The corresponding belief set $Bel^{se}(\Psi)$ is such that $Mod(Bel^{se}(\Psi)) = \{ \omega_0, \omega_2, \omega_8 ,  \omega_9 , \omega_{10} , \omega_{11}, \omega_{12} , \omega_{13}   \}$. Suppose we learn that {\it condition 3}
holds, namely we revise $\Psi$ by the propositional
formula $\mu = d$. According to the definition~\ref{defsemo} the revised epistemic state $\Psi \circ^{se}_{\triangleright}$ is represented by the partial pre-order on interpretations $\preceq_{\Psi \circ^{se}_{\triangleright}}$ graphically represented on Figure~\ref{exsemi}.

\begin{figure}[h]
  \caption{Representation of $\Psi\syntax^{se}\mu$}\label{exsemi}
  \vspace{3cm}
  \centering
  \rput(0,1.5){\rnode{I5}{$\omega_0=\omega_2=\omega_8=\omega_{10}=\omega_{12}$}}
  \rput(0,2.25){\rnode{I1}{$\omega_4=\omega_6$}}
  \rput(0,3){\rnode{I4}{$\omega_{14}$}}
  \rput(0,0){\rnode{I9}{$\omega_9=\omega_{11}=\omega_{13}$}}
  \rput(-2,0.75){\rnode{I6}{$\omega_1=\omega_3$}}
  \rput(0,0.75){\rnode{I7}{$\omega_5=\omega_7$}}
  \rput(2,0.75){\rnode{I8}{$\omega_{15}$}}
  \rput(-3,2.25){\rnode{L1}{$Mod(\lnot\mu)$}}
  \rput(-3,0.25){\rnode{L2}{$Mod(\mu)$}}
  \psline[linestyle=dotted](-4,1.125)(4,1.125)
  \ncline[nodesep=3pt]{<-}{I9}{I6}
  \ncline[nodesep=3pt]{<-}{I9}{I7}
  \ncline[nodesep=3pt]{<-}{I9}{I8}
  \ncline[nodesep=3pt]{<-}{I6}{I5}
  \ncline[nodesep=3pt]{<-}{I1}{I4}
  \ncline[nodesep=3pt]{<-}{I5}{I1}
  \ncline[nodesep=3pt]{<-}{I8}{I5}
  \ncline[nodesep=3pt]{<-}{I7}{I5}
\end{figure}

\noindent The belief set corresponding to $\Psi
\circ^{se}_{\triangleright} \mu$ is such that
$Mod(Bel^{se}(\Psi  \circ^{se}_{\triangleright} \mu)) = \{ \omega_9,
\omega_{11} , \omega_{13}  \}$ and since
$Mod(\mu) =\{\omega_1, \omega_3, \omega_5, \omega_7, \omega_9, \omega_{11}
, \omega_{13}, \omega_{15} \}$, it can be checked
that \\
$Mod(Bel^{se}(\Psi  \circ^{se}_{\triangleright} \mu)) = min(Mod(\mu),
\preceq_{\Psi} )$.

\end{example}

\subsubsection{Syntactic extension }

We now present the syntactic extension of this revision operation to
partial pre-orders.
Let $\Psi$ be an epistemic state, $\Psi$ is syntactically represented by a
partially ordered belief base,
denoted by $\preceq_{\Sigma}$, where $\Sigma$ is a set of propositional
formulas, and $\preceq_{\Sigma}$ is a partial
pre-order on $\Sigma$. The revision of $\preceq_{\Sigma}$ by a
propositional formula $\mu$ leads to a
 partially ordered belief base denoted by $\preceq_{\Sigma
\circ^{sy}_{\triangleright} \mu}$ as follows:

Let us denote by $U$ the set of the disjunctions between $\mu$ and the
formulas of $\Sigma$, more formally,
$U = \{ \phi \lor \mu,$ such that $\phi \in \Sigma$ and $\phi\lor\mu\not\equiv\top\}$.

\begin{definition}
Let $\Psi$ be an epistemic state, represented by a partially ordered belief
base $\preceq_{\Sigma}$, the revision
of $\Psi$ by $\mu$ leads to a revised epistemic state $\Psi
\circ^{sy}_{\triangleright} \mu$
represented by a partially ordered belief base where
$\Sigma \circ^{sy}_{\triangleright} \mu = \Sigma \cup U \cup \{ \mu \}$
and $\preceq_{\Sigma \circ^{sy}_{\triangleright} \mu}$ is such that:
\begin{itemize}
\item 
  $\forall\phi\lor\mu\in U:\;\phi\lor\mu\prec_{\Sigma\circ^{sy}_{\triangleright} \mu}\mu$,

\item 
  $\forall \phi \in \Sigma: \;  \mu\prec_{\Sigma\circ^{sy}_{\triangleright}  \mu} \phi$,

\item 
  $\forall \phi, \phi' \in \Sigma: \phi \preceq_{\Sigma} \phi'$
  iff $\phi \preceq_{\Sigma \circ^{sy}_{\triangleright} \mu}\phi'$,

\item 
    $\forall \phi, \phi' \in \Sigma: \phi \preceq_{\Sigma} \phi'$
    iff $ \phi\lor\mu \; \preceq_{\Sigma \circ^{sy}_{\triangleright}\mu} \;  \phi'\lor\mu$.
\end{itemize}
\end{definition}

\begin{example}
\label{ex6}

We come back to example~\ref{ex2}. Let $\Psi$ be the epistemic state, where
$\Sigma = \{ b \to a, b \land c \to \lnot a, d \to
a \}$, %we represent the epistemic
%state by a partial pre-order on $\Sigma$, denoted by 
$\preceq_{\Sigma}$ is such that it only contains one constraint:
%as follows:
$  b \land c \to \lnot a \prec_{\Sigma}  b \to a$. %and $d
%\to a$.

\noindent %Suppose we learn that {\it condition 3} holds, we 
Let us revise $\Psi$ by the propositional formula $\mu = d$. According to the definition of the revision operation
$\circ^{sy}_{\triangleright}$,
the revised epistemic state $\Psi \circ^{sy}_{\triangleright} \mu$ is represented by the following partial pre-order on Figure~\ref{exsyno}.

\begin{figure}[h]
  \caption{Representation of $\Psi \circ^{sy}_{\triangleright} \mu$}\label{exsyno}
\vspace{3cm}
  \centering
  \rput(0,0){\rnode{R6}{$(b\land c\to\lnot a)\lor d$}}
  \rput(0,0.75){\rnode{R5}{$b\to a\lor d$}}
  \rput(0,1.5){\rnode{R4}{$d$}}
  \rput(1,2.5){\rnode{R3}{$d\to a$}}
  \rput(-1,2.25){\rnode{R2}{$b\land c\to\lnot a$}}
  \rput(-1,3){\rnode{R1}{$b\to a$}}
  \ncline[nodesep=1pt]{<-}{R6}{R5}
  \ncline[nodesep=1pt]{<-}{R5}{R4}
  \ncline[nodesep=1pt]{<-}{R4}{R3}
  \ncline[nodesep=1pt]{<-}{R4}{R2}
  \ncline[nodesep=1pt]{<-}{R2}{R1} 
  \psline[linestyle=dotted](-3.5,1.875)(3.5,1.875)
  \psline[linestyle=dotted](-3.5,1.125)(3.5,1.125)
  \rput(-3,2.5){\rnode{R1}{$\Sigma$}}
  \rput(-3,1.5){\rnode{R2}{$\mu$}}
  \rput(-3,0.5){\rnode{R3}{$U$}}
\end{figure}

% $\Sigma \circ^{sy}_{\triangleright} \mu$ denoted by
%$\preceq_{\Sigma \circ^{sy}_{\triangleright} \mu}$ as follows:

%\noindent 
%$  
%( b \to a) \lor d
%\prec_{\Sigma \circ^{sy}_{\triangleright} \mu} 
%( b \land c \to\lnot a) \lor d
%\prec_{\Sigma \circ^{sy}_{\triangleright} \mu} d
%b \land c \to \lnot a \prec_{\Sigma\circ^{sy}_{\triangleright} \mu}  b \to a
%$

%\noindent and

%\noindent $
%(d \to a ) \lor d
%\prec_{\Sigma \circ^{sy}_{\triangleright} \mu} 
%d \to a \prec_{\Sigma \circ^{sy}_{\triangleright} \mu} d
%$.

\noindent Remark that since $(d \to a ) \lor d$ is a tautology, we do not take it into account.%the second part
%of the partial pre-order amounts to %$ d \prec_{\Sigma \circ^{sy}_{\triangleright} \mu}  d \to a$.

\end{example}

\subsection{Extension of possibilistic revision}

We now present the extension of the possibilistic revision to partial
pre-orders. In a possibility theory framework \cite{dlp1992} %this approach \cite{DuP92, DuP97}, 
 an epistemic state 
$\Psi$ is represented by a possibility distribution $\pi$. Each
interpretation is assigned with
a real number belonging to the interval $[ 0, 1 ]$. The value $1$ means
that the interpretation is totally possible,
whereas the value $0$ means that the interpretation is totally impossible. 
A possibility distribution induces a total pre-order $\leq_\pi$ on interpretations in the following way: $\omega\leq_\pi\omega'$ iff $\pi(\omega)\geq\pi(\omega')$. For more details on possibility theory and the revision of possibility distributions, see \cite{DuP92, DuP97}.

%A partial
%pre-order on interpretations is associated with possibility distribution $\pi$.

\subsubsection{Semantic extension }

Let $\Psi$  be an epistemic state represented by a partial pre-order $\leq_\Psi$. The possibilistic revision of $\Psi$ by
a propositional formula $\mu$ leads to a revised epistemic state $\Psi
\circ^{se}_{\pi} \mu$, represented by a partial pre-order on interpretations, denoted by $\preceq_{\Psi\circ^{se}_{\pi} \mu}$  which considers that all the counter-models of the new item of information $\mu$ as
impossible and preserves the relative
ordering between the models of $\mu$. More formally:

\begin{definition}\label{defposssem}
Let $\Psi$ be an epistemic state and $\mu$ be a propositional formula, the
revised epistemic state
$\Psi \circ^{se}_{\pi} \mu$ corresponds
to the following partial pre-order:
\begin{itemize}
\item if $\omega, \, \omega' \, \in Mod(\mu)$ then $\omega \preceq_{\Psi\circ^{se}_{\pi} \mu} \omega'$ iff $\omega \preceq_{\Psi} \omega'$,
\item if $\omega, \, \omega' \, \not\in Mod(\mu)$ then $\omega =_{\Psi\circ^{se}_{\pi} \mu} \omega'$,
\item if $\omega \in Mod(\mu)$ and $ \omega' \not\in Mod(\mu)$ then $\omega\prec_{\Psi \circ^{se}_{\pi} \mu} \omega'$.
\end{itemize}
\end{definition}

\noindent According to this definition it is easy to check that
$Mod(Bel^{se}(\Psi  \circ^{se}_{\pi} \mu)) = min(Mod(\mu), \preceq_{\Psi} )$.

\begin{example}

\noindent Let us consider again example~\ref{ex1}. 

We revise $\Psi$ by the propositional
formula $\mu = d$. According to the definition~\ref{defposssem}, the revised epistemic state $\Psi \circ^{se}_{\pi}\mu$ is represented by the following partial pre-order on interpretations $\preceq_{\Psi \circ^{se}_{\pi}\mu}$, graphically represented on Figure~\ref{exsemp}. \begin{figure}[h]
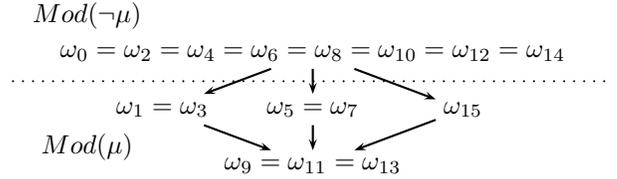

\caption{Representation of $\Psi\circ_\pi^{se}\mu$}\label{exsemp}
\vspace{2.5cm}
  \centering
    \rput(0,1.5){\rnode{I5}{$\omega_0=\omega_2=\omega_4=\omega_6=\omega_8=\omega_{10}=\omega_{12}=\omega_{14}$}}
    \rput(0,0){\rnode{I9}{$\omega_9=\omega_{11}=\omega_{13}$}}
    \rput(-2,0.75){\rnode{I6}{$\omega_1=\omega_3$}}
    \rput(0,0.75){\rnode{I7}{$\omega_5=\omega_7$}}
    \rput(2,0.75){\rnode{I8}{$\omega_{15}$}}
    \ncline[nodesep=3pt]{<-}{I9}{I6}
    \ncline[nodesep=3pt]{<-}{I9}{I7}
    \ncline[nodesep=3pt]{<-}{I9}{I8}
    \ncline[nodesep=3pt]{<-}{I8}{I5}
    \ncline[nodesep=3pt]{<-}{I7}{I5}
    \ncline[nodesep=3pt]{<-}{I6}{I5}
    \psline[linestyle=dotted](-4,1.125)(4,1.125)
    \rput(-3,2){\rnode{L1}{$Mod(\lnot\mu)$}}
    \rput(-3,0.25){\rnode{L2}{$Mod(\mu)$}}
\end{figure}

\noindent The belief set corresponding to $\Psi  \circ^{se}_{\pi} \mu$ is
such that $Mod(Bel^{se}(\Psi  \circ^{se}_{\pi} \mu)) = \{ \omega_9, \omega_{11} ,\omega_{13}  \}$.

\end{example}

\subsubsection{Syntactic extension }
We now present the syntactic extension of the possibilistic revision operation to partial pre-orders. Let $\Psi$ be an epistemic state, syntactically represented by a partially ordered belief base.  The revision of $\preceq_{\Sigma}$ by a propositional formula $\mu$ leads to a  partially ordered belief base denoted by $\preceq_{\Sigma \circ^{sy}_{\pi} \mu}$ as follows:

\begin{definition}\label{defposssyn}

The revision of $\Psi$ by $\mu$ leads to a revised epistemic state $\Psi\circ^{sy}_{\pi} \mu$ represented by a partially ordered belief base $\Sigma \circ^{sy}_{\pi} \mu = \Sigma \cup  \{ \mu \}$ where $\preceq_{\Sigma \circ^{sy}_{\pi} \mu}$ is such that:
\begin{itemize}
\item 
  $\forall \phi \in \Sigma$: $\quad \mu  \; \prec_{\Sigma\circ^{sy}_{\pi} \mu} \phi\; $,
\item 
  $\forall \phi, \phi' \in \Sigma$:  $ \phi \preceq_{\Sigma} \phi'$ iff $\phi \; \preceq_{\Sigma \circ^{sy}_{\pi} \mu} \;
\phi'$.
\end{itemize}
\end{definition}

\begin{example}

We come back to example \ref{ex2},  where $\Sigma = \{ b \to a, b \land c \to \lnot a, d \to a \}$, and $\preceq\Sigma$ only contains one constraint: $  b \land c \to \lnot a \prec_{\Sigma}  b \to a$.

\noindent
Let us revise $\Psi$
by the propositional
formula $\mu = d$. According to the definition~\ref{defposssyn},  xthe revised epistemic state $\Psi \circ^{sy}_{\pi} \mu$ is represented by
the following partial pre-order on
 $\Sigma \circ^{sy}_{\pi} \mu$ denoted by
$\preceq_{\Sigma \circ^{sy}_{\pi} \mu}$ and graphically represented by Figure~\ref{exsynp}.

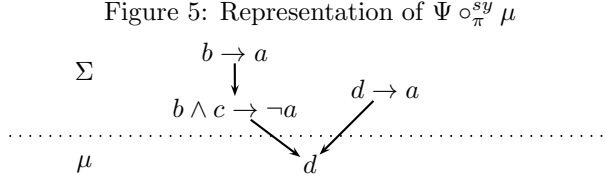
\begin{figure}[h]
  \caption{Representation of $\Psi \circ^{sy}_{\pi} \mu$}\label{exsynp}
\vspace{1.5cm}
  \centering
  \rput(0,0){\rnode{R4}{$d$}}
  \rput(1,1){\rnode{R3}{$d\to a$}}
  \rput(-1,0.75){\rnode{R2}{$b\land c\to\lnot a$}}
  \rput(-1,1.5){\rnode{R1}{$b\to a$}}
  \ncline[nodesep=1pt]{<-}{R4}{R3}
  \ncline[nodesep=1pt]{<-}{R4}{R2}
  \ncline[nodesep=1pt]{<-}{R2}{R1} 
  \psline[linestyle=dotted](-4,0.375)(4,0.375)
  \rput(-3,0){\rnode{R1}{$\mu$}}
  \rput(-3,1.25){\rnode{R2}{$\Sigma$}}
\end{figure}

%$d 
%\prec_{\Sigma \circ^{sy}_{\pi} \mu}  
%b\to a  
%\prec_{\Sigma \circ^{sy}_{\pi} \mu}   
%b \land c \to \lnot a$ 

%and $ d \prec_{\Sigma\circ^{sy}_{\pi} \mu} d \to a$

\end{example}

%We provide a generalization of the revision of an epistemic state
%represented by a total pre-order
%to the revision of an epistemic state represented by partial pre-order with
%two revision operators
%$\circ^{se}_{\triangleright}$ and $\circ^{se}_{\pi}$ at the
%semantic level and two revision operators $\circ^{sy}_{\triangleright}$ and
%$\circ^{sy}_{\pi}$ at the
%syntactic level. As mentioned in the introduction, in case of partial
%ignorance an agent only holds a
%partial pre-order on interpretations, since 
Note that revising a partial pre-order, with $\circ_\pi^{se}$ and $\circ_\triangleright^{se}$, 
carry away some incomparabilities. Hence, 
after a certain number of successive revisions the resulting partial
pre-order on interpretations
converges to a total pre-order on interpretations, more formally:

\begin{proposition}

Let $\preceq_{\Psi}$ be a partial pre-order on interpretations, there exists a
sequence of formulas $(\mu_1, \mu_2, \cdots , \mu_n)$
such that the resulting partial pre-order after successive revisions
\begin{itemize}
\item $(((\preceq_{\Psi}\circ^{se}_{\triangleright} \mu_1)
\circ^{se}_{\triangleright} \mu_2) \circ^{se}_{\triangleright} \cdots
\circ^{se}_{\triangleright} \mu_n)$ is a total pre-order, and
\item $(((\preceq_{\Psi}\circ^{se}_{\pi} \mu_1) \circ^{se}_{\pi} \mu_2)
\circ^{se}_{\pi} \cdots
\circ^{se}_{\pi} \mu_n)$ is a total pre-order.
\end{itemize}

\end{proposition}

The interest of such a result stems from the fact that starting from total
ignorance about a topic,
successive revisions lead to a partial pre-order on interpretations, and we now
know how to perform these revisions. Moreover, after a certain number of
revision the partial pre-order converges
to a total pre-order that can be revised according to the results
previously obtained in \cite{BDP99}, \cite{DuP92}.

%%%%%%%%%%%%%%%%%%%%%%%%%%%
\section{From syntax to semantics}
%%%%%%%%%%%%%%%%%%%%%%%%%%%
We now present the mapping from a partially ordered belief base
$\preceq_{\Sigma}$ to a partial pre-order on
interpretation $\preceq_{\Psi}$.

\begin{definition}
  Let $\Sigma$ be a partially ordered belief base and $\omega$ be an
  interpretation. We denote by $\lceil \omega, \Sigma \rceil$
  the set of preferred formulas of $\Sigma$ falsified by $\omega$. We define
  a partial pre-order on interpretations as follow:
    $$\omega \preceq_{\Psi,w} \omega' \textrm{ iff }  \lceil \omega', \Sigma \rceil \preceq_{\Sigma,w} \lceil \omega, \Sigma \rceil.$$% \textrm { or  }

\noindent Where $\preceq_{\Sigma,w}$ is given by definition~\ref{def1}
%    \lceil \omega', \Sigma \rceil = \lceil \omega, \Sigma \rceil.$$
\end{definition}

%%According to this definition, we define two mappings from
%$\preceq_{\Sigma}$ to $\preceq_{\Psi}$, namely
%$wt$ and $st$ using $\preceq_{\Sigma,w}$ and $\preceq_{\Sigma,s}$
%respectively, as partial pre-order on subsets of $\Sigma$.
%In the following we only use $\preceq_{\Sigma,s}$, but the same results
%hold for $\preceq_{\Sigma,w}$.

%According to this definition, we define a mapping $\preceq_{\Sigma}$ to $\preceq_{\Psi}$, denoted by $we()$, namely\footnote{In the following we only use $\preceq_{\Sigma,w}$, but the same results hold for $\preceq_{\Sigma,s}$, defined using strong preference.}:

\begin{example}
\label{ex9}
We come back to example \ref{ex2},% Let $\Psi$ be the epistemic state, 
 where $\Sigma = \{ b \to a, b\land c \to \lnot a, d\to a \}$, %we represent the epistemic
%state by a partial pre-order on $\Sigma$, denoted by $\preceq_{\Sigma}, $
 and $\preceq_{\Sigma}$ is defined as $  b\land c \to\lnot a\prec_{\Sigma} b \to a$.

\noindent The sets of preferred formulas of $\Sigma$ falsified by the
interpretations are the following:

\begin{center}
  \begin{tabular}{l}
    $\lceil\omega_0, \Sigma\rceil  = \lceil\omega_2, \Sigma\rceil= \lceil\omega_8, \Sigma\rceil = \lceil\omega_{9}, \Sigma\rceil =\emptyset$, \\
    $\lceil\omega_{10}, \Sigma\rceil =   \lceil\omega_{11}, \Sigma\rceil = \lceil\omega_{12}, \Sigma\rceil = \lceil\omega_{13}, \Sigma\rceil = \emptyset$,\\
    $\lceil\omega_1, \Sigma\rceil  = \lceil\omega_3, \Sigma\rceil = \{d\to a \}$,\\
    $\lceil\omega_4, \Sigma\rceil = \lceil\omega_6, \Sigma\rceil = \{ b \to a\}$,\\
    $\lceil\omega_5, \Sigma\rceil  = \lceil\omega_7, \Sigma\rceil = \{ b \to a, d \to a \}$,\\
    $\lceil\omega_{14}, \Sigma\rceil = \lceil\omega_{15}, \Sigma\rceil = \{ b \land c \to \lnot a \}$.\\
  \end{tabular}
\end{center}

\noindent According to definition of $\preceq_{\Sigma,w}$, it can be easily checked that the computation of $\preceq_{\Sigma,w}$ leads to the same partial pre-order than the
one used in the semantic representation of 
$\Psi$ in example \ref{ex1}, namely $\preceq_{\Sigma,w} = \preceq_{\Psi}$.
%and $st(\preceq_{\Sigma}) =  \preceq_{\Psi}$.

\end{example}

We are now able to establish the equivalence between the syntactic
representation of epistemic states by means of
partially ordered belief bases and the semantic representation of epistemic
states by means partial pre-orders on interpretations.

\begin{theorem}
Let $\Psi$ be an epistemic state represented, on one hand by a partially
ordered belief base $\preceq_{\Sigma}$ and
on the other hand by a partial pre-orders on interpretations
$\preceq_{\Psi}$. Let $\circ^{sy}_{\triangleright}$ and
$\circ^{se}_{\triangleright}$
be the syntactic and semantic revision operators stemming from the history
of the observations, and
$\circ^{sy}_{\pi} $ and $ \circ^{se}_{\pi}$
be the syntactic and semantic possibilistic revision operators. Let $we$ %$st$
%\footnote{The same result holds for the transformation $wt$ which
%uses $\preceq_{\Sigma,w}$ as partial pre-order on subsets of $\Sigma$.} 
 be the mapping from a partially
ordered belief base $\preceq_{\Sigma}$ to a partial pre-order on
interpretation $\preceq_{\Psi}$. The following result holds:

\begin{itemize}
\item 
  $we(\preceq_{\Sigma} \circ^{sy}_{\triangleright} \mu) = we(\preceq_{\Sigma} )  \circ^{se}_{\triangleright} \mu$,
\item 
  $we(\preceq_{\Sigma} \circ^{sy}_{\pi} \mu) = we(\preceq_{\Sigma} )\circ^{se}_{\pi} \mu$.
\end{itemize}

\end{theorem}

We illustrate this theorem by the following example.

\begin{example}

Let $\Psi$ be the epistemic state of example~\ref{ex2}, where
$\Sigma = \{ b \to a, b \land c \to \lnot a, d \to a \}$,  and $\preceq_{\Sigma}$ is such that: $  b \land c \to \lnot a \prec_{\Sigma}  b \to a$.

According to example \ref{ex6} the revised epistemic state $\Psi\circ^{sy}_{\triangleright} \mu$ is represented by the partial pre-order on $\Sigma \circ^{sy}_{\triangleright} \mu$ given by figure~\ref{exsyno}.

\noindent The sets of preferred formulas of $\Sigma\circ^{sy}_{\triangleright} \mu=\{ b \to a, b \land c \to \lnot a, d \to a, d, (b \to a)\lor d, (b \land c \to \lnot a)\lor d, (d \to a)\lor d   \}$ falsified by the interpretations are the following:\\

\begin{center}
  \begin{tabular}{l}
    $\lceil \omega_9, \Sigma \circ^{sy}_{\triangleright} \mu\rceil = \lceil \omega_{11}, \Sigma \circ^{sy}_{\triangleright} \mu\rceil = \lceil \omega_{13}, \Sigma \circ^{sy}_{\triangleright} \mu\rceil = \emptyset$,\\

    $ \lceil \omega_1, \Sigma \circ^{sy}_{\triangleright} \mu\rceil = \lceil \omega_3, \Sigma \circ^{sy}_{\triangleright} \mu\rceil = \{ d \to a \}$,\\

    $ \lceil \omega_5, \Sigma \circ^{sy}_{\triangleright} \mu\rceil = \lceil \omega_7, \Sigma \circ^{sy}_{\triangleright} \mu\rceil = \{ b\to a, d \to a \}$,\\
    
    $\lceil \omega_{15}, \Sigma \circ^{sy}_{\triangleright} \mu\rceil = \{ b \land c \to \lnot a \}$,\\
    
    $\lceil \omega_0, \Sigma \circ^{sy}_{\triangleright} \mu\rceil
    = \lceil \omega_2, \Sigma \circ^{sy}_{\triangleright} \mu\rceil
    = \lceil \omega_8, \Sigma \circ^{sy}_{\triangleright} \mu\rceil=$\\
    $= \lceil \omega_{10}, \Sigma \circ^{sy}_{\triangleright} \mu\rceil
    = \lceil \omega_{12}, \Sigma \circ^{sy}_{\triangleright} \mu\rceil
    = \{ d \}$,\\

    $ \lceil \omega_4, \Sigma \circ^{sy}_{\triangleright} \mu\rceil  =
    \lceil \omega_6, \Sigma \circ^{sy}_{\triangleright} \mu\rceil = \{  (b\to a)\lor d \}$,\\

    $\lceil \omega_{14}, \Sigma \circ^{sy}_{\triangleright} \mu\rceil
    = \{ (b \land c \to \lnot a)\lor d \}$.\\
  \end{tabular}
\end{center}

\noindent According to definition~\ref{defsemo} %of $\preceq_{\Psi\circ^{se}_{\triangleright} \mu,w} $, 
 it can be checked that $\preceq_{\Psi\circ^{se}_{\triangleright} \mu,w} $ is identical to the one given by figure \ref{exsyno}.

Since from example \ref{ex9} $we(\preceq_{\Sigma}) =  \preceq_{\Psi}$ we have
$we(\preceq_{\Sigma})  \circ^{se}_{\triangleright} \mu = \preceq_{\Psi\circ^{se}_{\triangleright} \mu}$ and
$we(\preceq_{\Sigma} \circ^{sy}_{\triangleright} \mu) =\preceq_{\Psi\circ^{se}_{\triangleright} \mu,s}$
it can be checked that $\preceq_{\Psi \circ^{se}_{\triangleright} \mu,s} =\preceq_{\Psi \circ^{se}_{\triangleright} \mu}$,
therefore $we(\preceq_{\Sigma} \circ^{sy}_{\triangleright} \mu) =we(\preceq_{\Sigma} )  \circ^{se}_{\triangleright} \mu$
and the equivalence between syntactic representation and semantic
representation holds.
\end{example}

%%%%%%%%%%%%%%%%%%%%%%%%%%%
\section{Syntactic computation of $Bel^{sy}(\Psi)$}
%%%%%%%%%%%%%%%%%%%%%%%%%%%

Let $\Psi$ be an epistemic state represented by a partially ordered belief
base $\preceq_{\Sigma}$, we now present the mapping from $\Sigma$ to $Bel^{sy}(\Psi)$. The computation of the corresponding belief set $Bel^{sy}(\Psi)$
is more complex and it is closely linked
to the ability of making syntactic inferences
from $\preceq_{\Sigma}$ in order to deduce the agent's current beliefs.
This involves the definition of a partial pre-order between
consistent subsets of $\Sigma$, denoted by ${\cal C}$ and we generate
a partial pre-order between
consistent subsets of $\Sigma$ %depending whether we use
using $\preceq_{\Sigma}$ described in definition \ref{def1}. or
%$\preceq_{\Sigma,s}$ described in definition \ref{def2}.

\begin{definition}
  Let $C, C' \in {\cal C}$:  $C \preceq_{{\cal C},w} C'$ iff $\{\phi': \phi' \not \in C' \}
  \preceq_{\Sigma,w} \{\phi: \phi \not \in C \}$,
%\begin{itemize}
%\item $ C \preceq_{{\cal C},w} C'$ iff $\{\phi': \phi' \not \in C' \}
%\preceq_{\Sigma,w} \{\phi: \phi \not \in C \}$,
%\item $ C \preceq_{{\cal C},s} C'$ iff $\{\phi': \phi' \not \in C' \}
%\preceq_{\Sigma,s} \{\phi: \phi \not \in C \}$.
%\end{itemize}
\end{definition}

Intuitively, $C$ is preferred to $C'$ if the best formulas outside $C$
are less preferred than the best formulas outside $C'$. We denote by $CONS_{w}(\Sigma)$ the set of
preferred consistent subsets of $\Sigma$ with respect
to $\preceq_{{\cal C},w}$, namely: %more formally:
%\begin{definition}
%Let ${\cal C}$ be the set of consistent subsets of $\Sigma$: 
$CONS_{w}(\Sigma)= min({\cal C}, \preceq_{{\cal C},w}).$
%$CONS_{s}= min({\cal C}, \preceq_{{\cal C},s})$.
%
%\end{definition}

\begin{example}
We come back to example \ref{ex2}, % Let $\Psi$ be the epistemic state, where
%$\Sigma = \{ b \to a, b \land c \to \lnot a, d \to
%a \}$, we represent the epistemic
%state by a partial pre-order on $\Sigma$, denoted by $\preceq_{\Sigma}, $
%as follows:
%$  b \land c \to \lnot a \prec_{\Sigma}  b \to a$ and $d
%\to a$.
the following array illustrates the set of consistent subsets of $\Sigma$:

$$
\begin{array}{|l|l|l|}
\hline
C_i & \{\phi \in C_i \} & Min (\{\phi \not \in C_i \},\preceq_\Sigma)\\
\hline
\hline
%C_0 & \emptyset & b \land c \to \lnot a, b \to a, d
%\to a \\
%C_1 & b \land c \to \lnot a &  b \to a, d \to a\\
%C_2 &  b \to a & b \land c \to \lnot a, d \to a\\
%C_3 & d \to a & b \land c \to \lnot a,  b \to a\\
%C_4 &  b \to a, b \land c \to \lnot a & d \to a\\
%C_5 & b \to a, d \to a  & b \land c \to \lnot a\\
%C_6 & b \land c \to \lnot a, d \to a  &  b \to a \\
%C_7 &  b \to a, b \land c \to \lnot a, d \to a &
%\emptyset\\
C_0 & \emptyset                    & b \land c \to \lnot a, d\to a \\
C_1 & b \land c \to \lnot a          &  b\to a, d \to a\\
C_2 &  b \to a               & b \land c \to \lnot a, d \to a\\
C_3 & d \to a                & b \land c \to \lnot a, \\
C_4 &  b \to a, b \land c \to \lnot a & d \to a\\
C_5 & b \to a, d \to a        & b \land c \to \lnot a\\
C_6 & b \land c \to \lnot a, d \to a  &  b \to a \\
C_7 &  b \to a, b \land c \to \lnot a, d \to a & \emptyset\\
\hline
\end{array}
$$

According to the definition of $\preceq_{{\cal C},w}$, we obtain the partial pre-order graphically represented by Figure~\ref{exci}. \begin{figure}[h]
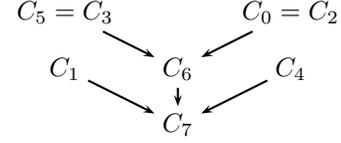

\caption{Representation of $\preceq_{\cal C},w$}\label{exci}
\vspace{2cm}
  \centering
    \rput(0,0){\rnode{C7}{$C_7$}}
    \rput(-1.5,0.75){\rnode{C1}{$C_1$}}
    \rput(1.5,0.75){\rnode{C4}{$C_4$}}
    \rput(0,0.75){\rnode{C6}{$C_6$}}
    \rput(-1.5,1.5){\rnode{C5}{$C_5=C_3$}}    
    \rput(1.5,1.5){\rnode{C2}{$C_0=C_2$}}

    \ncline[nodesep=3pt]{<-}{C7}{C1}
    \ncline[nodesep=3pt]{<-}{C7}{C4}
    \ncline[nodesep=3pt]{<-}{C7}{C6}
%    \ncline[nodesep=3pt]{<-}{C6}{C0}
    \ncline[nodesep=3pt]{<-}{C6}{C2}
    \ncline[nodesep=3pt]{<-}{C6}{C5}

\end{figure}

\end{example}

%The syntactic inference is defined as follows:

%\begin{definition}
%Let $\phi$ be a propositional formula,\\
%$\phi$ is syntactically inferred from $\preceq_{\Sigma}$ iff $\forall C \in
%CONS_{s}$, $C \cup \lnot \phi$ is inconsistent.
%\end{definition}

The syntactic definition of $Bel$ is: 
$$ Bel^{se}[\Psi]=\bigvee_{C\in Cons_w(\Sigma)}C.$$

This means that the syntactic inference can be defined as: $\phi$ is inferred from $\preceq_\Sigma$ iff $\forall C\in CONS_{s}$, $C \cup \lnot \phi$ is inconsistent.

%Using this definition the following result can be established:

\begin{theorem}
Let $\Psi$ be the epistemic state, let $\circ^{sy}_{\triangleright}$ and
$\circ^{se}_{\triangleright}$
be the syntactic and semantic revision operators stemming from the history
of observations, and  $\circ^{sy}_{\pi} $ and $ \circ^{se}_{\pi}$
be the syntactic and semantic possibilistic revision operators. The
following result holds:

\begin{itemize}
\item $Bel^{sy}(\Psi \circ^{sy}_{\triangleright} \mu)  \equiv Bel^{se}(
\Psi \circ^{se}_{\triangleright} \mu)$,
\item $Bel^{sy}(\Psi \circ^{sy}_{\pi} \mu)  \equiv Bel^{se}( \Psi
\circ^{se}_{\pi} \mu)$.
\end{itemize}
\end{theorem}

We illustrate this theorem with an example:

\begin{example}

Let $\Psi$ be the epistemic state defined in example \ref{ex5} where the revision
by $\mu = d$ leads, %as detailed in example
%\ref{ex5}, 
to the belief set $Bel^{se}(\Psi  \circ^{se}_{\triangleright}
\mu)$ such that $Mod(Bel^{se}(\Psi  \circ^{se}_{\triangleright} \mu)) = \{
\omega_9, \omega_{11} , \omega_{13}  \}$.
%since $Mod(Bel^{se}(\Psi  \circ^{se}_{\triangleright} \mu)) = min(Mod(\mu),
%\preceq_{\Psi} )$.\\

%$Bel^{sy}(\Psi \circ^{sy}_{\triangleright} \mu) $ is computed as follows.

On the syntactic level, we can check that, since $\Sigma\cup\{\mu\}\cup U$ is consistent, that the preferred elements for $\preceq_{\mathcal{C},w}$, $Cons_w(\Sigma\circ_\triangleright^{sy}\mu)$ is simply the subset composed of all elements of $\Sigma\circ_\triangleright^{sy}\mu$. Namely we have 
$Cons_w(\Sigma_{\circ_\triangleright^{sy}\mu})= \{ \{ b \to a,b \land c \to \lnot a,d \to a,d,$ $(b \to a)\lor s,$ $(b \land c \to \lnot a)\lor d,$ $(d \to a)\lor d   \}\}$ and hence $Mod(Cons_w)=\{\omega_9,\omega_{11},\omega_{13}\}$, as it is excepted.

% the set of all elements of $\Sigma_{\circ_\triangleright^{sy}\mu}$ is preferred to all other subset of $\Sigma_{\circ_\triangleright^{sy}\mu}$, we have so $Cons_w(\Sigma_{\circ_\triangleright^{sy}\mu})=\{ b \to a,b \land c \to \lnot a,d \to a,d,(b \to a)\lor s,(b \land c \to \lnot a)\lor d,(d \to a)\lor d   \}$.

%We have $Mod(Cons_w)=\{\omega_9=\omega_{11}=\omega_{13}\}$, which is the excepted result.

\end{example}

%%%%%%%%%%%%%%%%%%%%%%%%%%%
\section{Concluding discussion}
%%%%%%%%%%%%%%%%%%%%%%%%%

Since in certain situations an agent faces incomplete information and has
to deal with partially ordered information,
this paper proposed a semantic representation of an epistemic state by a
partial pre-order on interpretations as well as
a syntactic representation by a partially ordered belief base. The
extension to partial pre-orders of two revision
strategies already defined for total pre-orders are presented, and the
equivalence between the representations is shown.
We showed that after a certain number of successive revisions the partial
pre-order convergences to a total pre-order.

In a future work, we have to investigate the properties of these revision
operators and the presented approach could be generalized to the revision
of a partial pre-order by a partial pre-order, generalizing the approach
proposed in \cite{BKPP00}. Moreover, in order to
provide reversibility the encoding by polynomials of partial pre-orders on
interpretations and partially ordered belief base
could be investigated.

%In order to provide an efficient computation of the presented revision
%operators an 
Another future work is to develop algorithms for computing the belief set
at the syntactical case, and to apply them in
geographical information systems where available
information is often partially ordered.

%%%%%%%%%%%%%%%%%%%%%%%%%
\section {Acknowledgments}
%%%%%%%%%%%%%%%%%%%%%%%%%
This work was supported by European Community with the REV!GIS project
$\sharp$ IST-$1999$-$14189$.
http://www.cmi.uni-mrs.fr/REVIGIS

\bibliography{nmr02}

%\bibliography{biblio}

\end{document}